\documentclass{article}
\usepackage{graphicx} 
\usepackage{url}
\usepackage{color,soul}
\usepackage{authblk}
\usepackage{acl}
\usepackage{tikz}

\definecolor{cbblue}{HTML}{1F78B4}
\definecolor{cborange}{HTML}{FF7F00}
\definecolor{cbgreen}{HTML}{33A02C}

\newcommand{\nbEC}[1]{{\color{blue} \footnotesize #1}}

\newcommand{\addRN}[1]{{\color{red}#1}}

\title{No Such Thing as a General Learner: \\Language models and their dual optimization}

\author[1]{Emmanuel Chemla}
\author[2,3]{Ryan M. Nefdt}
\affil[1]{PSL University, EHESS, LSCP, CNRS, ENS}
\affil[2]{University of Cape Town}
\affil[3]{University of Bristol}
\date{-}

\begin{document}

\maketitle
\thispagestyle{plain}
\pagestyle{plain}


\begin{abstract}

    What role can the otherwise successful Large Language Models (LLMs) play in the understanding of human cognition, and in particular in terms of informing language acquisition debates?
    To contribute to this question, we first argue that neither humans nor LLMs are general learners, in a variety of senses. We make a novel case for how in particular LLMs follow a dual-optimization process: they are optimized during their training (which is typically compared to language acquisition), and modern LLMs have also been selected, through a process akin to natural selection in a species. From this perspective, we argue that the performance of LLMs, whether similar or dissimilar to that of humans, does not weigh easily on important debates about the importance of human cognitive biases for language.

\end{abstract}

\section{Introduction}

There has been a wealth of collaborative work recently on questions pertaining to whether language models display human-level understanding \cite{bender-koller-2020,ohmer2024forms} or have access to semantic meaning or reference \cite{mandelkern2024,baggio2024,lederman2024}. The question thus arises as to whether or not LLMs could be interpreted as scientific theories of language---of language processing, or of language acquisition---, and it receives both strongly positive support \cite{piantadosi2023} as well as largely negative critique \cite{Katzir2023,kodner2023,FoxKatzir:2024}. 
At the heart of the matter is the issue of learning, i.e.~what kind of learners are LLMs, and how does this compare to the learners that humans are.

We offer novel insight on the matter, first by discussing in section~\ref{sec:general} the pervasive assumption concerning the possibility of a general learner, and what that means exactly. We repeat that under a strong interpretation of the concept, there is simply no such thing at all, since a learner is first and foremost defined by necessary learning preferences (or biases), and we argue that under weaker definitions LLMs cannot count as general learners either, given their own evolutionary-like maturation (section~\ref{sec:LLMgeneral}). 
In section \ref{sec:benchmark}, we discuss the consequences of this for the current field that is structured around benchmarks mostly concerned with measures of the final, trained states of LLMs. 
In section~\ref{sec:imposs}, we apply our arguments to the evaluations more focused on the learning stages of LLMs. One debate asks whether LLMs are not too powerful, often phrases around the question as to whether `impossible' languages, that allegedly cannot be learned by humans, can be learned by LLMs. We add to the debate the fact that, even when trained to learn \emph{possible} languages, parts of the languages that LLMs learn are indeed impossible. This shows that the biases of LLMs are different from ours, and remind us that an adequate model of learning has to learn neither too little nor too much. 

\section{What is a general learner?}\label{sec:general}

In the history of philosophy, there has been a vigorous debate between rationalists and empiricists on the nature of concept acquisition, which dates back to at least Locke and Leibniz (see \citealp{Cowie1999}). In a more contemporary cognitive scientific setting, this discussion has played out in terms of the contrast between learning via innate mechanisms versus learning via general principles of experience. Rationalism (which favours the former) still commands much of the landscape in linguistics \cite{Chomsky1966} with some connections to issues in machine learning \cite{Buckner2024}. From these philosophical corners comes the idea of a general learner or a blank slate (\textit{tabula rasa}).\footnote{Again, see \citet{Cowie1999} for nuance on the interpretations of both Leibniz and Locke on concepts.} In the contemporary setting, there are several salient interpretations of this concept: 
\begin{enumerate}
    \item General learners as unbiased or unconstrained or \textit{un}primed towards specific languages,
    \item[1'] General learners who may acquire any language, given sufficient training,
    \item General learners as \textit{vanilla} learners whose generality allows them to be picked out off-the-shelf without too much consideration, 
    \item General learners who acquire language in a domain-general and/or non-modular manner. 
\end{enumerate}

We know now that (1) does not exist in any real sense. In a learning setting, the training corpus under-determines what the full target is: the target language may be restricted to the training corpus, or it may contain other sentences. The task of a learner is to deploy its biases, whatever they are, to make guesses about what, beyond the exemplars it has seen, may or may not belong to the language. (As \citet[34]{Pinker2004} put it `the mind cannot be a blank slate, because blank slates don't do anything'). 
(1') is a more subtle version of (1). It can be well captured through the work of Gold \cite{gold1967language}, providing results under which training can lead to approach any language infinitely well. Still, results there concern potentially infinite training, approximate learning, and a restricted set of languages, which will be key in later discussions.
(2) will be of central interest to us. It may be an implicit temptation in some work in NLP, but we will claim that it is not applicable to modern LLMs. And (3) is often assumed to be a property of artificial learners by default, which would by design learn homogeneously across domains whatever they are fed, while human learners, some claim, learn language differently than they way in which they learn in other domains.


\section{Are LLMs general learners?}
\label{sec:LLMgeneral}

The question then becomes whether LLMs can be considered general learners under any reasonable interpretation of the term. We argue that the answer is no.

\subsection{No such thing as a general learner} In terms of (1), it is known that LLMs are not unbiased or `unconstrained' learners, simply because there is no such thing as a general learner. 
In Bayesian terms: every type of priors, including flat priors, are priors. 
\citet{kodner2023} for instance stress this point by citing \citet{kharitonov2021}, showing that different model architectures make different structural choices based on ambiguous data sets.\footnote{Specifically, Long Short-Term Memory models (LSTMs) with attention and Transformers seem to infer hierarchical structure while LSTMs without attention and Convolutional Neural Networks (CNNs) opt for linear based representations on the same training sample which could have been generated by either.} 
What makes LLMs powerful then is the very fact that they are specialized. If they were not specialized, they would not make inductive leaps, and they would at best be able to reproduce their training data.\footnote{Both the issues of interpolation versus extrapolation \cite{HASSON2020} and the possibility/extent of data contamination are an open questions in NLP research. We will bracket them for present purposes.} And \citet{piantadosi2023} equally denies the existence of unconstrained or `anything goes' models: `it’s important to realize that there is no such thing as an “anything goes” model, in that any model will necessarily have certain tendencies and biases'. The problem is a mathematical one: with little data, the set of possible options is not much constrained, so that powerful learners are those which would go beyond their data via opinionated learning, i.e.~thanks to their biases.

\subsection{Vanilla is a flavour too} Let us consider a second-order version of the question then, and ask whether the biases of LLMs can themselves be taken to be a default. It is a harder notion to define, so let us make it operational. If one were to randomly pick some learner, would it learn languages like us? Our point here is a novel one. We claim that such an idea of an off-the-shelf vanilla model that we can pick, as in (2) above, is equally questionable, and especially inapplicable in the contemporary context of LLMs. 

\begin{figure}
    \centering
    \includegraphics[width=\linewidth,trim=2.9cm 2cm 3cm 2cm,clip]{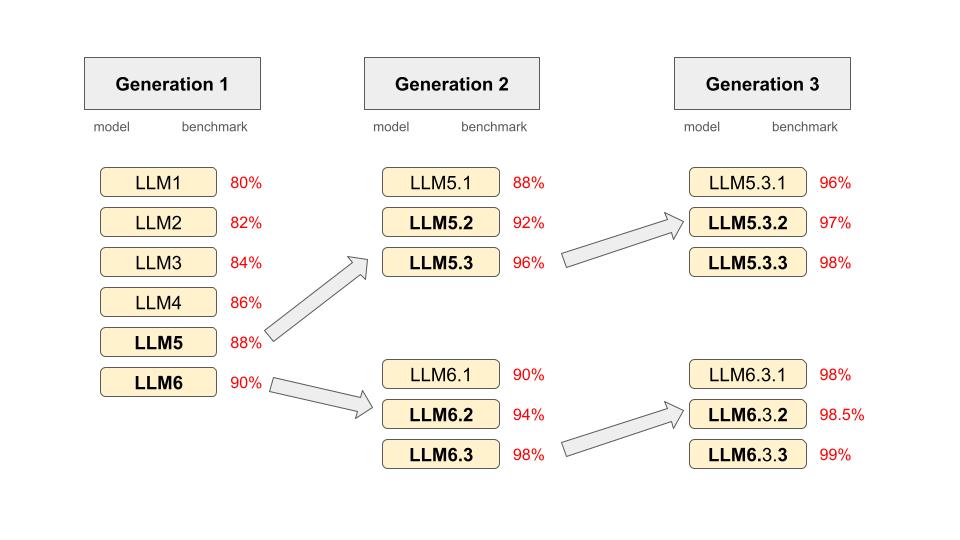}
    \caption{LLMs have followed a long chain of optimization, which have made them increasingly specialized for language learning.}
    \label{fig:generation-optimization}
\end{figure}

Thousands of labs and startups around the world have invested billions of dollars to test dozens or hundreds of LLMs per year for the last few decades. This amounts to millions of LLMs tested. Typically, as in a natural evolution process, a model would survive if it satisfies the company goals better. And similarly in academia: the standard format of results in the domain is a table, whereby models are compared according to established benchmarks. The model that surpasses the others (traditionally highlighted with boldface) on most of these benchmarks is naturally the one that receives most attention from the community, and attracts follow-up studies, see Fig.~\ref{fig:generation-optimization}. In other words, models are evaluated and compared based on some (manually crafted) fitness measures, and those that survive get to be improved further in the next generations of studies. 
Hence, the models that have survived today, BERT, Llama, GPT-4, LaMDA, are not default or random samples of a diverse population; they are the models selected for their performance (or fitness), further optimized to perform even better, while others have been selected out for their lack of relative accuracy. 
This is a full-blown optimization process. It happens through small and large changes (e.g., small changes in the number of layers of a model, vs.\ more significant innovations to move to LSTM, transformer or CNN architectures). Overall, the space of models that have been explored is very vast, even in terms of regular, unintelligent evolution, and this landscape has been explored through a recognizable optimization method. 


Hence, modern LLMs are engineering devices, optimized for specific tasks, and their evolution-like optimization has been accelerated by the human hand for their specific goals (more on the variety of possible goals in the next section).%
\footnote{This, of course, does not mean that they have evolved or been designed in the same way as human learners have. Human/animal evolution is a bottom-up blind process. Although, see \citet{Dennett2017} for a similar evolutionary account of top-down intelligent systems like LLMs.}
In fact, \citet{yax2024phylolminferringphylogeny} can be seen as pushing this metaphor to its extreme, as they propose a way to construct phylogenetic trees of LLMs (primarily in terms of `traits', that is their behavior as to what tokens they generate and how they fare with respect to various benchmarks, rather than direct structural or historical considerations). 

Returning to cognitive science, modern LLMs cannot count as vanilla, off-the-shelf learners. The fact that a given model ends up capturing facts about natural language can thus no longer be used against a nativist approach to language learning, since the given model was optimized for that purpose (or a closely related one, see below). The fact that \emph{some} learner designed for the task ends up learning language will not help measure the depth of the biases that humans have for learning language. To bear on the nativism debate, one would need to agree on a process to randomly sample learners among learners with no specific biases for language (a notion to be defined in the first place), and show that these learners nonetheless learn language like humans do, from the same input that we use. Doing so is an arduous task, and we are aware of no attempt to tackle it. 
Hence, LLMs do not provide a good bed to dismiss arguments from `the poverty of the stimulus', a series of arguments relating the quantity and quality of information that a learner can adduce from the input it is given vs.\ from its biases. One reason which has been put forward in the past is that LLMs receive a very different input (certainly a much larger one). Another reason that we push forward is that their biases have been tailored for the task.

\subsection{General as domain general} The last possible interpretation of `general learner' corresponds to a distinction in cognitive science between domain-specific and domain-general. Domain specificity is often associated with modularity and cognitive impenetrability \cite{pylyshyn1984,Fodor2015}. The issue is whether, in humans at least, language is a specific module that is cognitively sealed from connection with general or central mechanisms such as planning, memory, reasoning, etc.\footnote{This doesn't mean that language is not utilised in those other systems.} For an artificial model to be a domain specific learner in this sense, it would have to learn other, non-linguistic datasets in a different way, perhaps with different parameters or structure. Pre-transformer natural language processing (NLP) might have mirrored some of these cognitive distinctions, i.e.~different architectures for different kinds of cognitive tasks. Recurrent Neural Networks (RNNs) were largely used for language tasks, CNNs for vision, Visual Geometry Group (VGGish) for audio classification, etc. However, with the advent of transformers and multi-head attention \cite{vaswani2023} the ideal of a general architecture for all kinds of tasks was seemingly within reach.\footnote{This characterisation is different from standard accounts in grammar induction in important respects. For example, \citet{Yang2022} make use of simple putatively domain-general operations (like their $pair(L,C)$ which concatenates character C to list (or string) L or $append(X,Y)$ which appends lists) to train models to induce the most likely formal grammar in terms of Bayesian conditional probabilities. However, unlike in corpus studies or LLM training, the generating grammar follows specific formal rules for construction of the dataset (or stringset) which the inducing grammar has to infer. Thus the formal languages might already display characterisations of linguistic structure unless shown to characterise non-linguistic patterns.} This picture is misleading for two reasons. The first is that blended architectures have never gone out of fashion, and CNNs have made a comeback for language tasks. The choice between different architectures remains a choice about what works best for rather specific goals, trading-off on other tasks: pick the model that performs the best for your purposes. 

\begin{figure}
    \centering
    \includegraphics[width=\linewidth,trim=5.5cm 5cm 0cm 5cm,clip]{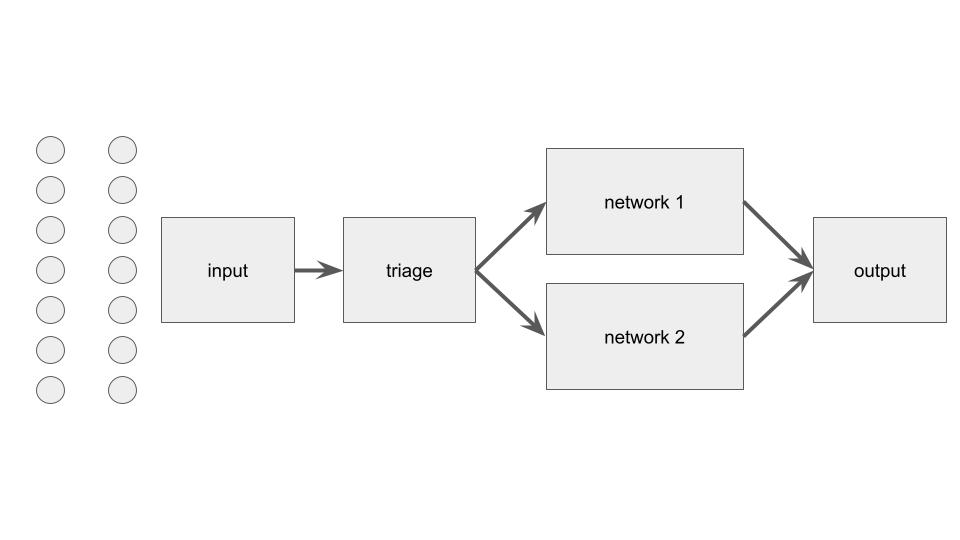}
    \caption{In large artificial networks, it is very possible that different tasks are treated by different sub-parts of the models, including largely independent parts. The existence of a single model to treat several tasks is therefore not an argument against some modularity between these tasks.}
    \label{fig:no-modularity}
\end{figure}

Secondly, even in the scenario in which distinct processes would be active for different tasks, this may be hidden in large artificial networks. The reason is as follows (see Fig.~\ref{fig:no-modularity}). Suppose that a network$_{1}$ is best to perform task$_{1}$ from input of type 1, and network$_{2}$ is best to perform task$_{2}$ from input of type 2. Now, network$_{1}$ and network$_{2}$ can be concatenated into a large network$_{1+2}$. Modern networks are so large that they may have space to contain both networks, and keep them completely separate for all relevant computational purposes. All that would be needed then to construct a single network that performs tasks 1 and 2 appropriately is a (general purpose) mechanism that can distinguish the two types of inputs and redirect them to the right subnetwork (see \citealp{Pfeiffer2023ModularDL}). In other words, the existence of a network that can grow into being able to process two tasks at the same time, is not evidence that the two tasks are performed in the same way. Such considerations unfortunately trivialize the notion of modularity. There is currently very little argument that can be made to say that there is not a separation between a network$_1$ and a network$_2$ within a single LLMs performing two tasks. The reason is mostly the same as in humans: it is hard to produce anatomical or behavioral arguments for the existence or absence of modules in a brain or in an artificial network.

Thus, despite the temptation, there is no obvious sense to be made to the idea that LLMs are general learners of language. As we will now see in more detail, this reduces the impact that a comparison between humans and LLMs can have and how this can inform debates from linguistics and cognitive sciences.

\section{Are LLMs human-like? The role of benchmarks}
\label{sec:benchmark}

LLMs learn something about human languages. What does this mean? Operationally, it means that they pass tests that have been designed to challenge earlier models. In the domain, these tests are the golden standard, they are called benchmarks. The development of these benchmarks itself is regarded as an important step forward, since benchmarks reveal interesting phenomena on which models can be tested and compared, and they provide new engineering targets when no model satisfyingly pass them.

These benchmarks are numerous and cover simple phenomena as well as some that are regarded as very sophisticated by the experts. Some benchmarks are born informally, either as the result of practitioners inspecting their results and detecting new oddities to be corrected or, now that some of these models are exposed to the general public, emerging from the exploration of a huge workforce of people trying to find the limits of these systems and pointing them out. 

\subsection{The nature of the benchmarks}
\label{sec:benchmark-competence}

While these benchmarks typically start out as new challenges and boundaries, boundaries are pushed by new progress, and in turn become the testimony of new achievements. The prime output of LLMs is probabilities of words or phrases, and therefore it is also the prime source of metric for benchmarks. In the domain of syntax, for instance, benchmarks are typically about whether a model attributes a higher probability to sentences which are regarded as grammatical, compared to minimally different sentences which are ungrammatical. For this purpose, the work of linguists centered on creating scientifically rich minimal pairs, is of paramount importance (see \citealp{sprouse2012assessing} for discussion about the reliability of minimal pairs, and \citealp{Chemla2023} for their scientific role, among many other work, including disagreements). 
In the context of LLMs, some have forcefully disputed the typical metrics used for success, as metrics that do not at all make an interesting comparison with human behavior (see, e.g., \citealp{Lan-LLM-PoStim}).
For now, let us nonetheless assume that these benchmarks are perfect. We will now argue that they are not external observers of the game. They are either measures of progress made or measures of progress to be made. And this has important consequences about how LLMs can and cannot be used in debates of interest to cognitive science. 

Taking a concrete and notable example, it has been shown that some LLMs pass tests correlated with the apparent knowledge of constraints on \textit{wh}-movement \cite{wilcox-etal-2018,Linzen2021}, phenomena that are considered complex, in the sense that they may involve hidden structure, thought to be hard to acquire from surface form or performance data alone. While such phenomena used to be a learning challenge indeed, some now consider it as a proof of success of LLMs. And this success is interpreted as proving that the technology is going in the right direction, which is motivation to continue in that direction. By similar measures, models have been shown to fail at some directly comparable phenomena (e.g., parasitic gaps or across-the-board movements, \citealp{Lan-LLM-PoStim}). One may take it as a serious failure, or one may react more constructively and consider that it provides for the next interesting landmark to pass, and that it will eventually be passed by improved models. This positive attitude contains a very important twist: new models are designed to pass more benchmarks, they are as such \emph{optimized} for new goals. Let us unfold this.

\subsection{Benchmarks as implicit objective functions}

There are two ways in which benchmarks can be used. The first way is to use benchmarks to put models to the test, to compare models. This first, natural use of benchmarks separates them from the optimization process, which in turn means that we can see models trained on a given objective (typically the next word prediction) and  pass many other challenges without prior exposure to them. This suggests that all learning paths lead to the learning of a whole language, that is a language which would be very well determined by its surface form, and with enough signal about its deep structure, that goes above and beyond what one could think should come out from the original training task.

Now consider another way to use these benchmarks: they could be integrated explicitly in the training objective: models could be explicitly trained to (also) pass these challenges. If we were to do so, these tests could not be said to come out for free, that their solution can be read out from the surface form of the linguistic material a learner may be exposed to. Now, as said before, an LLM is the result of a \emph{dual} optimization; they are not vanilla learners.\footnote{And yet we ignore the further optimization process of fine-tuning via reinforced learning by human feedback (RLHF) and other such procedures.} First, they are optimized during training, on an explicit loss, typically through gradient descent (that is tiny, accumulative modifications). Second, they also follow an evolution-like optimization process, through which the best models are selected. In that latter part of the optimization process, benchmarks do play a role. That latter part happens very transparently in the domain when models are ranked based on their results across benchmarks. And the process also occurs under the hood prior to publication or in start-ups, when models exposed to the scientific community or shipped to the industry are fine-tuned by their designers to pass these benchmarks. In the context of psychology, this would count as an instance of a publication bias for positive results, which is considered one of the source of the replication crisis \cite{lindsay2015replication}. There is nothing problematic to this in a short-term engineering context, since the end goal is to find the most efficient model, but it has scientific consequences.

Overall, benchmarks, even when they are praised for their arbitrator status, that is to compare models that exist at a particular point in time, end up playing a part in the \emph{selection} of models, that is, in the optimization of the surviving models.
So, when a new benchmark emerges, whether a formal or an informal one, it naturally becomes part of the optimization process, implicitly (through the selection process, before or after publication/distribution) or explicitly (if added formally to the training objective).

\subsection{Benchmarks about behavior vs benchmarks about learning}
\label{sec:benchmark-learning}

Despite the explosion of benchmarks, not all relevant aspects of behavior are really covered. Importantly in the context of understanding learners, development itself is rarely tested. The order of acquisition of different phenomena is relatively stable across human languages and humans \cite{kuhl2004early,dupoux2018cognitive}, but we know little about the order in which LLMs acquire the different linguistic phenomena. This is partly because it is not clear what would count as a timeline for acquisition for LLMs. A `younger' LLM may be one that has been exposed to a smaller training corpus, or to a simpler training corpus (how to define complexity without circularity?), one that has been run for fewer training epochs, one that navigates its environment with a purpose and can be claimed to have passed milestones along the way (\citealp{carta2023groundinglargelanguagemodels}), etc. The training corpora that are currently used are implausibly large in the first place \cite{frank2023bridging}. Some (e.g., \citealp{baroni2022proper}) rightfully argue that mere size is not a fair comparison, because an LLM may need more text to compensate the absence of other types of information (they do not have vision, or ability to interact with their environment). Or they may need more input to compensate evolution times, although then one asks whether LLMs are good models of humanity (generations of humans), not of humans. In any event, then, the notion of time of learning \textit{ex hypothesi} is different in an LLM and in humans. If the time scale is different, the order of acquisition may also be organized different within that time frame. Overall, it is not easy to decide how to use the course of an LLM training and compare it to that of human language learning. 

Yet, there are rare and interesting studies to consider. For instance \citet{evanson2023} (see also \citealp{lavechin2022statistical}) suggest that there is significant overlap in the stages of learning between a human child and an LLM.
Within that first picture of the situation, LLMs conform to the order of acquisition in children. For example, they learn subject-verb agreement in declarative sentences before they master it in questions, or embedded clauses. One may like to use such work to address the empiricism versus nativism debate in language acquisition: how much of the human behavior is driven by the data they are exposed to (empiricism) versus their priors (nativism). Again, given that LLMs are not vanilla learners, their learning is constrained both by their extensive training data, and by the biases that have been grown into them. The fact that they end up following a human-like order of development, may either be reason to believe that there is something natural to this order, i.e.~naturally readable from the training data (empiricism), or that the appropriate `naturalness' has been forged into them (nativism).

\begin{figure}
    \centering
    \includegraphics[width=\linewidth,trim=3cm 6.5cm 3cm 6.5cm,clip]{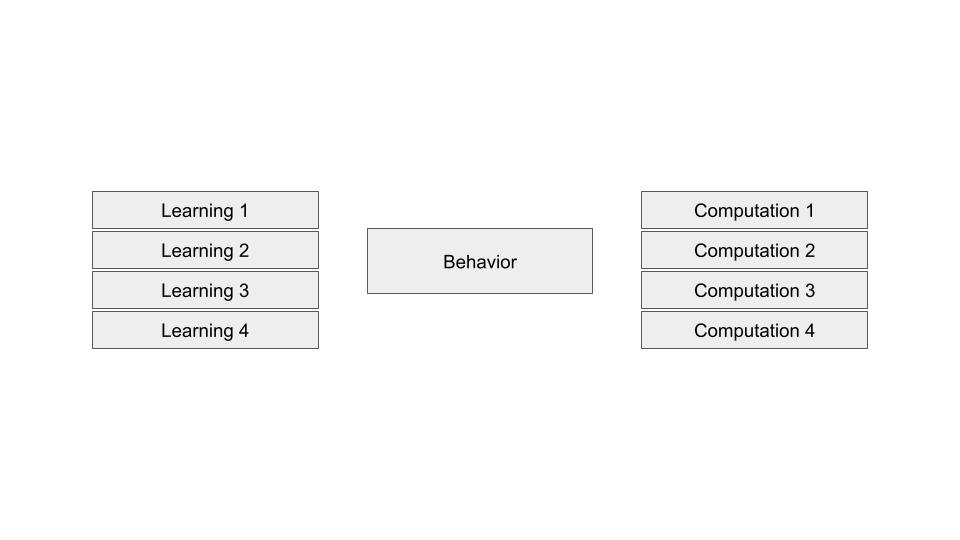}
    \caption{Several learning strategies and several computational models can lead to the same behavior, that is, a good fit of behavior still largely underspecifies a model of cognition. 
    }
    \label{fig:one-to-many}
\end{figure}

In sum, there is a one-to-many mapping between a set of behaviors that have been learned, and the learning strategies and path that could lead to learning them (see Fig.~\ref{fig:one-to-many}). As a result, we need more assessments of the learning part of the LLMs, beside assessments of their final behaviors. Once we have those, however, a delicate balance will appear again: if we optimize our systems, directly or indirectly, to follow the target developmental path, then we will lose the ability to argue that this developmental path is natural. We will raise this issue again in section~\ref{sec:imposs}, where some studies have moved to `controlled' languages specifically designed to test learning biases in ecological as well as non-ecological situations, the latter of which presumably not being the target of engineering efforts.

\subsection{Behavior vs representations and computations}

Finally, we note that behavioral benchmarks, no matter how strict and exhaustive they may be, have another fundamental blind-spot. \citet{marr1982vision}'s legacy makes it clear that there is another important one-to-many mapping to be conscious of, one between behavior and underlying possible computations that could lead to this behavior. For this reason, to test whether LLMs are good models of human processing (more than acquisition, for reasons discussed above), we can test alignments between brain activations and LLM activations \cite{Pasquiou2022NeuralLM}. Allegedly, a good alignment at this level would provide more evidence that the underlying representations and computations are similar. A pessimistic remark however is that there has to be some fit at this level: if behavior is the same, then some features will be represented by both systems. The degree of fit that is needed at the anatomical level to claim equivalence of computations is a challenging question. Across humans, there is some alignment between how different brains activate while parsing language. The alignment is not perfect, if only because the anatomy is different there. It is thus hard to know what type and extent of alignment we should be seeking between a brain and a neural network to evaluate their alignment at the computational level (see discussion in \cite{jalouzot2024metriclearningencodingmodelsidentify}).

\section{Are LLMs human-like? The case of impossible languages}
\label{sec:imposs}

Benchmarks are typically designed to measure whether LLMs and humans are as successful as each other on a particular behavior after training. Here we introduce another way in which humans and LLMs have been compared. This avenue is directly concerned with learning (see section~\ref{sec:benchmark-learning}), though still mostly directed at the final outcome of learning. It is also one that does not strive for matching some positive performance, but for matching the \emph{in}ability to learn certain languages compared to others. Accordingly, it is not constrained to study the learning of actual, human languages, but can study the learning of languages that do not exist in practice, and allegedly so for principled reasons. 

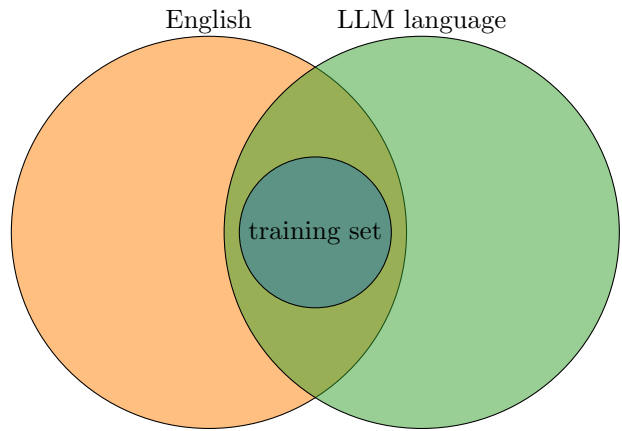
\begin{figure}[!ht]
    \centering
\begin{tikzpicture}
    \draw[fill=cborange, fill opacity=0.5] (-1.4,0) circle (2.6) node[below=2.2cm] {};
    \draw[fill=cbgreen, fill opacity=0.5] (1.4,0) circle (2.6) node[below=2.2cm] {};
    \draw[fill=cbblue, fill opacity=0.5] (0,0) circle (1) node[below=1.2cm] {};
    
    \node at (0,0) {training set};
    \node at (-1.4,2.8) {English};
    \node at (1.4,2.8) {LLM language};
\end{tikzpicture}

    \caption{From a given input, different learners will learn different languages. The typical remark is that a defective LLM may learn a \emph{subset} of English (the left part will be missing). But this left part has been decreasing. Now we are also finding that LLMs also learn languages of their own. These languages could be languages that humans will never learn, even if given the relevant input. Or they could be extensions of the languages that humans would learn say with English, and these later cases are in fact not only attested and obviously very foreign to humans, but there existence is a necessary, mathematical consequence of the form of these models.
    }
    \label{fig:enter-label}
\end{figure}

\subsection{The classic impossible language debate}\label{sec:impossible languages}

At the centre of the recent controversy as to the learning capabilities of humans compared to language models is the idea that LLMs are insensitive to a pivotal distinction between possible and impossible languages \cite{MORO2023}. As human learners, on the other hand, we are defined by our limitations. 

The idea of impossible languages stems from the possibility of identifying basic building blocks of human language or Universal Grammar (UG). As \citet[40]{Chomsky1991} described it `knowing something about UG, we can readily design languages that will be unattainable by the language faculty'. Once the fundamental principles are determined, the stimulus-dependent parameters can be set to different degrees for the realization of individual languages or reset to create impossible configurations (cf.~the Principles and Parameters framework). 
A less stringent version of this framework is to say that humans have strong biases against languages following these configurations. Such biases, as we now have discussed, necessarily exist for a learner, and the question as to whether these biases are strict or can be overruled is, in fact, a rather secondary question that should better remain open.

Empirically, one of the earliest experiments designed to test this hypothesis involved not a machine but a polyglot savant named Christopher \citep{SMITH1993}.\footnote{Christopher's apparent abilities, on the performance level, resided in his command and sensitivity to the complex morphology of a large number of languages.} The researchers exposed Christopher to two new languages, the North African language Berber, and an invented language with `impossible' elements called Epun. They attributed his linguistic prowess to an ability to reset parameters, something neurotypical language learners are supposed to be unable to do after a critical period. This implies that Christopher could technically keep acquiring languages well after puberty and thus be a perfect test subject for what is humanly acquirable and what is not, i.e.~what a human possible language is. They defined impossibility itself as either the contravention of some principle of the then current theory of generative grammar (Government and Binding) or an unattested construction in any known language. They expected Christopher to falter on Epun given his hypothesized exceptional language module and his relatively impaired general or central functions.

The results were complicated but they paved the way for later work on the topic. For example, neither Christopher nor their control groups were able to learn structure-independent operations while Christopher also reportedly struggled with structure-dependent but linguistically impossible constructions such as verb initial negative sentences lacking in negative morphemes. 

\citet{Moro2016} and his earlier experimental work upon which it is based, take the notion of structure dependence as a mark of the possible.\footnote{He also considers hierarchical constituency and recursion to be essential.} This means that elements of sentences are connected or dependent on other non-adjacent elements (or not necessarily adjacent) in the overall structure. In fact, impossibility is defined as the absence of this structure: `an impossible language is one in which dependencies can be rigidly determined by the position of a word in a linear sequence' \cite[30]{Moro2016}. The experimental research goes beyond behavioral tests to neuroimaging techniques testing whether the \textit{pars triangularis} in the Broca area (associated with syntactic processing) was activated in the same way when subjects were
taught structure-dependent rules versus ‘rules with rigid dependencies based on
the position of words in the linear sequence’ or ‘impossible rules’ \cite[55]{Moro2016}. Subjects were exposed to two sets of rules in a foreign language, one based on structure-dependence and the other on linear order, then placed within an fMRI. The resulting images displayed an augmentation of blood in the desired brain region for the structure rules and a diminished amount for the application of the linear rules. There is a lot to unpack and challenge here.\footnote{See \citet{Nefdt2024} for some discussion of the formal properties combined (and possibly conflated) in the experimental design. } But let this serve as some background to the claims concerning LLMs to which we now move.

\subsection{LLMs and impossible languages}

The aforementioned ideas can be applied to the present context. 
The claim is that LLMs process possible and impossible languages indistinguishably \cite{Watumull2023,MORO2023}. One \emph{a priori} reason to push that charge is that there is nothing visible in the way LLMs calculate what they calculate that could seem to create some biases that humans may have: in a transformer for instance, words are input from a set of positions, and nothing suggests that this could create a bias in favor of a hierarchical organization, rather than creating relative or absolute position preferences, and one may even be able to produce a formal proof that languages and backward languages could be treated equivalently (p.c.~Dominique Sportiche, see \citealp{Lan:LLMTheory2024} for empirical tests and discussions about the relevance of the debate).
For a review of some possibilities in this vein, see \citet{strobl-etal-2024-formal}. Nevertheless, the claim, in essence, is that LLMs are general learners insensitive to the structure of human language. In \citet[84]{MORO2023}'s words, `machines appear to be able to compute all sorts of impossible languages, including those based on ``flat'', i.e.~non-hierarchical rules'. It is unclear what `compute' means here as neither the experiments in \citet{SMITH1993} nor \citet{Moro2016} rule out learning linear rules by induction. An LLM could learn or generalize human possible languages via a different mechanism or path to impossible ones (see section~\ref{sec:LLMgeneral}). They go on to strengthen the claim to `LLMs and other types of transformer models learn impossible grammars just as well as human grammars' \cite[84]{MORO2023}. It is this point that \citet{kallini2024} directly challenge with their experiments. Here, our earlier distinction between general learners of type (1) and (1') becomes important, as it allows one to say that a learner may be able to learn any language, but possibly with more difficulty, typically computed in terms of input size.


Concretely, \citet{kallini2024} propose a test for the claim that LLMs do not distinguish between possible and impossible grammars. First, they construct a continuum of languages based on their degree of `impossibility'. 
On the impossible side of the spectrum are random word shuffles, the midpoint contains unnatural word orders or reversed string languages, and the possible end comprises the hierarchically structured (attested) languages.
These sets of languages are created by defining `perturbation functions' on sets of English sentences. The process involves both considerations of information theory (e.g., entropy rates) and formal language theory (e.g., the nested complexity of formal languages in the Chomsky Hierarchy). The relevance of their choice of languages has been disputed, but we will not address this worry: without such a decision it is impossible to evaluate the claims of \citet{MORO2023}, so it is methodologically necessary. They test the impossible language posit (that LLMs don't have our limits) using GPT-small models trained on a modified BabyLM dataset \cite{Radford2019}. 

In all three experiments they run, they find that LLMs trained on possible languages prefer human-like solutions, i.e.~natural grammar rules, to ungrammatical or impossible constructions. 
LLMs are thus deemed a reasonable comparison class from which we can learn about human linguistic cognition. Their results dovetail with our analysis as to the nonexistence of general learners. In fact, it should not be surprising that LLMs designed to learn possible languages prefer such structures to generally unattested impossible or unnatural ones. They are not general or vanilla learners. In this way, GPT-2 is not a default choice of model (as \citealp{kallini2024} note in terms of its information locality bias). 

It would, of course, be a significant result to show that certain LLMs show a preference for human natural language constructions over gerrymandered artificial ones. This is especially relevant given the claims of \citet{MORO2023,Watumull2023} and others skeptical of the role LLMs can play in linguistics or cognitive science. But the impact of such studies is attenuated in light of the fact that LLMs are not general learners of language and have never been so. In other words, the fact that LLMs learn what we do when primed on the kinds of data we are exposed to is interesting, but not conclusive of their status as cognitive scientific tools either. At best, they are Frankenstein versions of us, designed by men to look like us at 90\% and up. This is significant as an engineering device. But scientifically, it does not follow, conversely, that human language acquisition is a default, vanilla process, deprived of rich biases, arbitrary preferences, etc.


Results and research like \citet{kallini2024} (as well as \citealp{evanson2023} discussed in section~\ref{sec:benchmark-learning}) might seem to advocate for the use of LLMs in the comparative cognitive science of language. But they can also go in the opposite direction. 
If indeed we are correct that LLMs have undergone a process similar to natural selection in their competitive task-engineer design phases then their closeness to humans on particular tasks cannot be taken to prove any general purpose claims about innateness or nature versus nature. Neither their failure on particular tasks nor their success is determinate of their relevance without further qualification. If anything, the interpretation of the situation may be that it takes some efforts to obtain a learner with biases similar to humans.

\subsection{The secret language of LLMs (autoprompt)}

In the previous section, we discussed the ability or inability of humans and machines to learn languages manually designed to be in or out of scope of human languages, as delimited by theories of human languages.
Here we highlight the less discussed fact that LLMs may be learning \emph{more} than human languages indeed, using a different set of facts: LLMs may do so even when exposed to a proper training from actual human language.

For LLMs, learning is judged successful when the language can be used productively. It has been found that, as a side effect of learning say English, LLMs may learn a secret language of their own. It works as follows. LLMs are evaluated for being able to provide an accurate completion to a sentence such as `The capital city of [X] is ...', with X a variety countries. It has been shown however that if one is looking for the best way to ask that question to some LLMs, one may better use a different `prompt' than that English prompt (e.g., \citealp{wallace-etal-2019-universal, shin-etal-2020-AutoPrompt, zhong2021factualprobingmasklearning} and much work since then). Instead, there is some combination of token embeddings such that `w1 w2 w3 [X] ...' would be completed more often than the English prompt with the correct capital city. And the sequence of `words' `w1, w2, w3', which is called an autoprompt, really is nonsense to speakers of English or to any human.%
\footnote{One may argue that it is almost by design that such sequences could be found. English prompts are not perfect for LLMs, and Autoprompts are just taking advantage of this room for improvement. Autoprompts are formal optimization to fill in this room for improvement. And Autoprompts can even be created outside of the domain of words altogether. To LLMs, English words are just a finite number of points in a continuous high dimensional space. But Autoprompts can be created outside of these English word points, then not making any visible sense to us at all. And \cite{khashabi-etal-2022-prompt} showed that LLMs may have `translations' of any English prompt, close to any other English prompt. This shows that the universe of LLMs is much more vast than that of English, even when they are tasked to learn English.}

What is important here is that there are sentences understood by LLMs that do not look to us as a natural extension of what is in the input, which is English. While learning English, LLMs learn a secret language, an impossible language. As illustrated in Figure~\ref{fig:enter-label}, LLMs learn more than what was in the input. That much, we already expected would be the case, because that is what learners do: a learner \emph{generalizes} from the input it sees. Machines and humans, as learners, both extrapolate from their training input. What is striking here is the massive departure as to where LLMs and humans take this extrapolation. 
When exposed to a language like English, a young human learns English (or some subset thereof). They do not add an additional language or code the way LLMs do. Particular non-human biases seem to push LLMs in this direction.

Overall, LLMs learning from a corpus of a given human language would, thanks to their well-aligned biases, learn that language to a significant extent, no matter what learning means in that context. What we highlight here is that they do not only learn that language, they learn a much larger language that is completely out of the realm of anything human-like. Hence creating a language of their own, which would count as an impossible language.

\section{Conclusions}

LLMs mark a remarkable human achievement. However, the role they play in the theory of language learning and cognitive science in general is still much contested. For example, we have shown that their existence does not establish the possibility of a general learner. The way their engineering successes weigh in fundamental debates related to linguistic competence, language acquisition, and the limits of human cognition must be assessed with care in light of their history and engineering goals. LLMs are learners, learners with biases, as all learners. Their biases have been and continue to be sculpted efficiently. They follow a meta-optimization process led by insightful or serendipitous innovations, benefiting from top-down as well as bottom-up signals, that are injected explicitly or implicitly into them.

Models of human language learning and processing are made to match human behavior (engineering models are even made to surpass it). Such models help us in daily tasks, or can serve to understand these behaviors. But the mere existence of these models, now or in the future, does not refute the fact that human linguistic behavior is a result of evolution, which means that some aspects are likely to be the arbitrary result of historical accidents. Mimicking human behavior is not reproducing it, nor understanding it, yet. LLMs clearly have some role to play in the understanding of human language, and perhaps other cognitive systems, but properly harnessing these potential insights needs to take the evolution and selection of the models into consideration in order to be successful in the future. 

\bibliography{main-2}

\end{document}